\pdfoutput=1
\documentclass[11pt]{article}
\usepackage[margin=1in]{geometry}
\usepackage{rawfonts}
\usepackage{lscape}
\usepackage[table,dvipsnames]{xcolor}
\usepackage{caption}
\usepackage{subcaption}
\usepackage{amsmath}
\usepackage{algorithm}
\usepackage[noEnd=false,italicComments=true]{algpseudocodex}

\DeclareMathOperator*{\argmax}{arg\,max}

\newcommand{\colt}{\textsc{CoLT}}

\newcommand{\ace}{\textsc{ACE}}

\newcommand{\bw}{\cellcolor{Goldenrod!25}}
\newcommand{\bwf}{\cellcolor{Goldenrod!70}}
\newcommand{\bwo}{\cellcolor{Goldenrod!25}}

\newcommand{\hc}{\cellcolor{Gray!20}}

\newcommand{\bc}{\cellcolor{Cyan!25}}
\newcommand{\bcf}{\cellcolor{Cyan!70}}

\begin{document}

\title{Adapting to Teammates in a Cooperative Language Game}

\author{ Christopher Archibald \\ archibald@cs.byu.edu \\       
Computer Science Department\\
       Brigham Young University \\
       Provo, UT 84606 USA
\and Spencer 
  Brosnahan \\ sbrosnah@byu.edu  \\
       Computer Science Department\\
       Brigham Young University \\
       Provo, UT 84606 USA
}

\date{}

\maketitle

\begin{abstract}
The game of Codenames has recently emerged as a domain of interest for intelligent agent design. 
The game is unique due to the way that language and coordination between teammates play important roles.
Previous approaches to designing agents for this game have utilized a single internal language model to determine action choices. 
This often leads to good performance with some teammates and inferior performance with other teammates, as the agent cannot adapt to any specific teammate. 
In this paper we present the first adaptive agent for playing Codenames. 
We adopt an ensemble approach with the goal of determining, during the course of interacting with a specific teammate, which of our internal expert agents, each potentially with its own language model, is the best match. 
One difficulty faced in this approach is the lack of a single numerical metric that accurately captures the performance of a Codenames team. 
Prior Codenames research has utilized a handful of different metrics to evaluate agent teams. 
We propose a novel single metric to evaluate the performance of a Codenames team, whether playing a single team (solitaire) game, or a competitive game against another team. 
We then present and analyze an ensemble agent which selects an internal expert on each turn in order to maximize this proposed metric. 
Experimental analysis shows that this ensemble approach adapts to individual teammates and often performs nearly as well as the best internal expert with a teammate.
Crucially, this success does not depend on any previous knowledge about the teammates, the ensemble agents, or their compatibility. 
This research represents an important step to making language-based agents for cooperative language settings like Codenames more adaptable to individual teammates.
\end{abstract}

\section{Introduction} \label{sec:intro}
There are many settings in which it is important for artificial intelligence (AI) agents to utilize language to communicate with other agents, be they artificial or human. 
When coordinating with other agents using language it is important to have a certain amount of linguistic flexibility, so that no matter how another agent utilizes certain words, successful communication can occur. 
This can especially be important in \emph{ad-hoc coordination} \cite{stone2010ad}, when the identity of teammate agents is unknown at the time an agent is designed.
Choosing a single static language model that will lead to success with a wide variety potential teammates is a very challenging problem. 

In this article we focus on an adaptive approach to this problem of natural language coordination. 
We propose an AI agent that can adapt its use of language to achieve better performance with its current teammate, whoever that happens to be. 
We explore this idea within the specific setting of a popular language game: Codenames. 
In this game natural language is used by teammates to communicate. 
The state of the game, which is the focus of player communication, is also composed of natural language words. 
A key characteristic of a successful team is the compatibility between their understanding of language, particularly how teammates each view the connections and relationships between different words. 
The proposed adaptive agent uses the feedback obtained during gameplay to adjust which language model or representation it is using. 
This ability to adapt to an individual teammate without prior knowledge is an important asset in many domains where agents can be paired with a diverse set of previously unknown teammates and the goal is to maximize performance with each of them. 

\subsection{Codenames}
\emph{Codenames} \cite{codenames}, won the prestigious Spiel des Jahres award for the best board game of the year in 2015. 
The game of Codenames involves two teams (red and blue) playing on a board of 25 words.  
Each team consists of at least two players: one is the team's \emph{spymaster} and the others are \emph{guessers}\footnote{For simplicity we will refer to a single guesser on each team}. 
The 25 board words each belong to one of four categories. 
9 cards that belong to the team that goes first, while 8 belong to the other team. 
Additionally, one card is an \emph{assassin} card, and the remaining 7 are designated as \emph{bystanders}. 
The category to which each card belongs is known only to the two spymasters. 

The teams alternate turns, with each turn consisting of a clue given by the spymaster to his team's guesser and their subsequent guessing. 
Each clue $c = (w,n)$ consists of a single word $w$ and a number $n$, which generally indicates the number of board words that the spymaster is relating with the word clue $w$. 
The clue word $w$ cannot be any of the unguessed words on the board.
The guesser guesses one board word at a time and that card's category is revealed. 
The guessing continues until the guesser fails to guess a word belonging to their team, the guesser chooses to stop, or the guesser has made $n+1$ guesses\footnote{One exception to this is that if $n=0$, there is no limit to the number of correct guesses}. 
If the assassin word is ever guessed the guessing team loses immediately. 
Otherwise, the winning team is the first team to have all of their board words revealed. 
This means that if the red team guesses a blue word, that gets the blue team one step closer to victory. 

\subsection{Why Codenames?}
In the time since 2015, Codenames has attracted interest from AI researchers as a fascinating testbed in which to work. 
The game has many features that make it especially interesting to researchers in many fields.
It fundamentally involves language, creativity, and words, as well as unique combinations of cooperation and competition.

Language tasks inspired by, and very similar to, Codenames have been used as ways of exploring and evaluating semantic memory models to explain human word-association datasets in the fields of cognitive science and computer science \cite{kumar2021semantica,shen2018comparing}.
Codenames presents an interesting, natural, and useful way to determine compatibility of two language models.
Especially for humans, where we might not have an explicit language model, we can ask: ``How well does a human play Codenames with an AI based on this language model?'' and a model that does better with humans on this task could be considered more compatible with that human than another language model whose AI performed more poorly. 
Similarly, several researchers focused on computational creativity have identified Codenames as an ideal testbed to evaluate creativity in humans and AIs \cite{zunjani2019towards}, and to develop and evaluate AI language components that can form important building blocks for larger efforts towards computational creativity \cite{spendlove2022competitive}.

\begin{figure}
    \centering
    \includegraphics[width=0.2\textwidth]{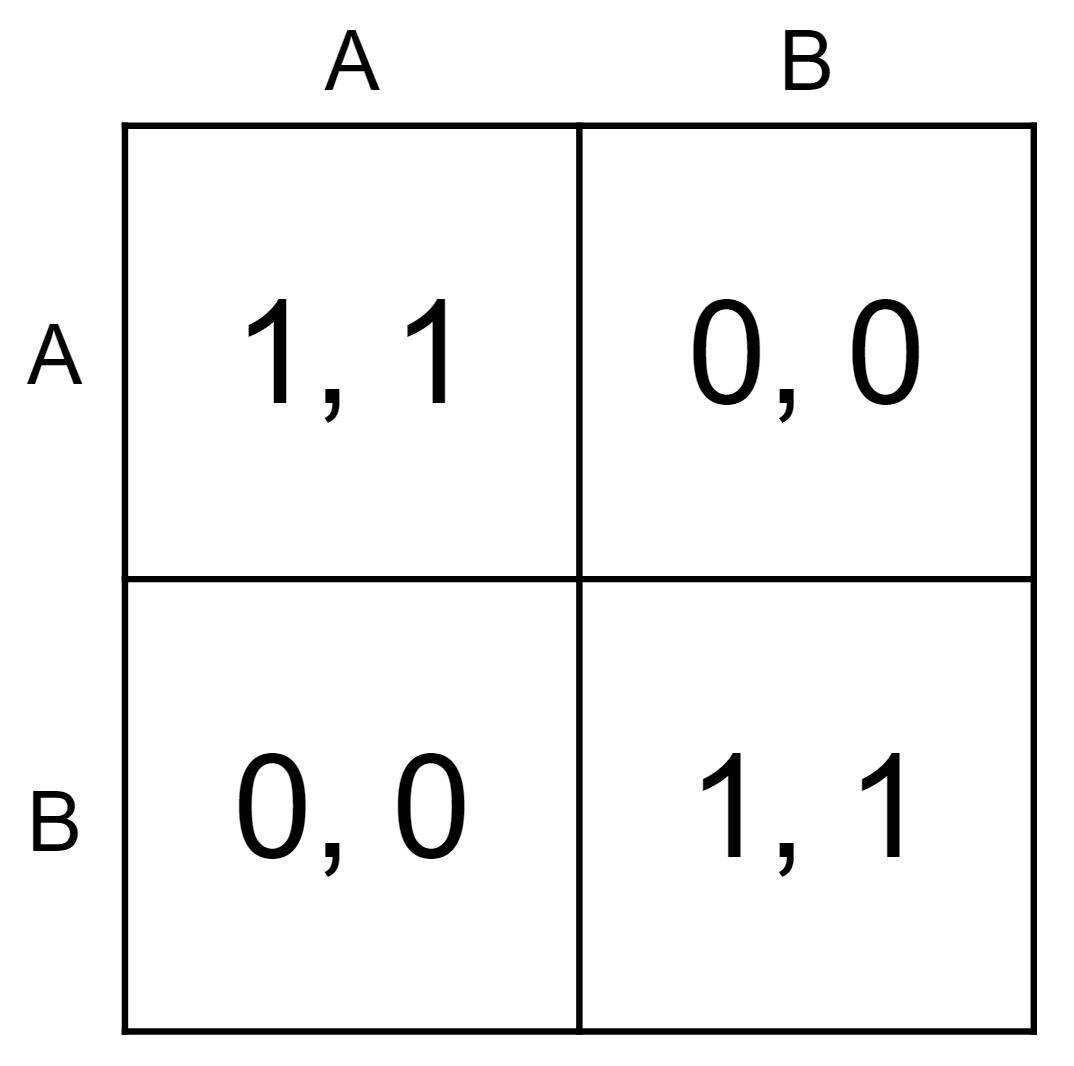}
    \caption{Coordination Game}
    \label{fig:coordination-game}
\end{figure}

Looking beyond the linguistic aspects of the game, Codenames also involves two main levels of strategy.
First is the interaction between the spymaster and the guesser on a single team. 
This interaction is one of pure coordination. 
In the game theory literature, the most simple game of pure coordination is called the \emph{Coordination Game} \cite{shoham2009multiagent}, shown in Figure \ref{fig:coordination-game}. 
In this game two players each have two possible actions. 
If both players select the same action, then they each get a reward of 1, while if the agents select different actions, they get a reward of 0.
Neither of the two actions is inherently better than the other, it only matters that both agents have selected the same one. 

We think of the coordination between teammates in Codenames in the same manner, although the space of possible $(w, n)$ actions is much larger in Codenames. 
In a certain sense, no word clue the spymaster could select to refer to a given set of board words is inherently better than any other.  
The important thing is that both spymaster and guesser have the same concepts of linguistic similarity of the words. 
In other words, it is important that they are each using compatible internal models of language.
This idea inspires our goal of designing an agent that adapts its choice of language model to most effectively coordinate with its current teammate instead of using a single language model.

The competitive portion of Codenames is strategically interesting in its own right. 
It consists of two teams trying to balance risk and reward in order to be the first to achieve the winning state. 
In a general sense, clues that have a higher $n$ are more risky, as they require the spymaster to associate more board words with a single clue word, and the guesser must be able to determine which board words were intended.
The most conservative approach is for a spymaster to give a clue that corresponds to only a single board word. 
This reduces the chances of an error (turning over an undesirable board card) but will require more turns to correctly identify all of the teams cards. 
A more risky strategy might try to frequently give clues for 3 or more words simultaneously. 
This increases the chances of an error, but when successful will require fewer turns to reveal all of the team cards, which will lead to a greater chance of beating the other team. 
The amount of risk a team should be willing to take on is, of course, going to depend on the context of the state of the game and what the board words are.
For example, if the other team only has one board card left, a spymaster should be more willing to give a clue for as many board words as are left, since there is a high probability the other team will successfully complete the game on the next turn, and so the team needs to win on this turn. 
Two existing metrics, which will be discussed in section \ref{sec:previous-codenames}, have focused roughly on these two aspects of a team's success: error rate and number of turns required.
One of the contributions of this work, described in section \ref{sec:score}, is a novel metric which approaches this tradeoff in a different, principled way. 

\subsection{Previous Codenames AI Research} \label{sec:previous-codenames}
Codenames was first introduced to the AI community in 2019 \cite{kim2019cooperation}. 
In that paper the first framework for both spymaster and guesser Codenames agents was presented. 
Several different language models were used within this framework, including WordNet \cite{miller1995wordnet}, word2vec \cite{mikolov2013efficient}, and GloVe \cite{pennington2014glove}. 
Each spymaster and guesser was evaluated based on how well they worked with a variety of teammates, including both teammates using the same language model as well as teammates with different internal language models. 
It was found that a concatenation of the word2vec and GloVe models worked best, and the authors suggested that other language models, like BERT \cite{devlin2018bert} and ELMO \cite{peters2018deep}, be explored in the future to see how well they work at the task. 

That paper, as well as subsequent work, evaluated teammate pairings in a solitaire version of Codenames, where the board is the same, but only one team gives clues and reveals cards. 
The game ends in either a win (all of the team's board cards successfully located) or a loss (the assassin or all of the other team's cards revealed).
A handful of metrics were used to evaluate the success of these teammate pairings. 
However, only two were subsequently used in almost every other paper on Codenames. 
The first was \emph{win rate}, the percentage of games a team won out of their games played. 
The second was \emph{win time}, which is the number of turns it took for the team to win games. 
The win time metric is computed looking only at the games the team won.
In Section \ref{sec:score} we discuss the shortcomings of using these two metrics and propose a single metric for evaluating Codenames teams.

Later work \cite{jaramillo2020word} explored other language models that could be used within the same basic Codenames agent framework, including GPT-2 \cite{radford2019language}. 
They found that the agent utilizing the GPT-2 based model outperformed the concatenated word2vec + GloVe model. 
Other work \cite{koyyalagunta2021playing} introduced BERT \cite{devlin2018bert} and BabelNet \cite{navigli2012babelnet} based AI approaches.
However, the main goal of this work was to improve the ability of AI-generated clues to be effective when used with human teammates. 
They showed that their DETECT method for scoring clues was able to improve the precision and recall of human agents in identifying which words were intended by specific AI-generated clues.  

One thing that has been true for all of the previously-discussed AI agents is that internally they utilize a single language model. 
Most of the work has focused on generating Codenames clues and guesses from a single language model and determining which language model is the best to use.
Additionally, all of the previous agent designs only consider the current board state when generating a clue or guesses, ignoring the results of previous turns. 
This memoryless strategy, combined with using a single language model, can lead to interesting suboptimal behavior by these computational agents. 
As one example, if a spymaster agents gives a clue and an incorrect board word is guessed, on the next turn the agent can give the exact same clue once again.
This occurs when the clue is still considered the best, given the static language model and memoryless clue-determining algorithm. 
Secondly, if the language model being used by an agent generates clues that are not compatible with the way their teammate guesser thinks about words, then they are simply out of luck, and will perform badly with that teammate forever. 

\subsection{Our Contributions}
In this article we make two main contributions. 
\begin{enumerate}
\item We present the \emph{Adaptive Codenames Ensemble} (\ace{}), the first adaptive AI agent framework for Codenames. 
The agent utilizes feedback from each turn during Codenames games to improve performance with a teammate over time. 
\ace{} internally contains a set of base Codenames agents, each of which can use a different language model and/or method for determining clues or guesses. 
We will refer to this set of internal Codenames agents as the set of \emph{experts}. 
The main challenge of this approach is to determine how to utilize the results of previous turns to select the expert to use on the next turn.
The \ace{} approach works for both clue-giving and guessing. 

\item We propose the \emph{Codenames Linear Team} (\colt{}) rating function, a novel method for evaluating team performance in the game of Codenames. 
This rating function is utilized internally by the \ace{} framework to track the success of each expert.
However, it can also be used on its own to evaluate the performance of agent teams in Codenames, either replacing or in addition to the win rate and win time metrics previously used in the literature.
As a single metric it has value in allowing for a direct ranking and comparison between different Codenames teams, whereas utilizing the existing two metrics does not always yield a definitive ordering.
\end{enumerate}

The remainder of this article proceeds as follows:  Section \ref{sec:overview} gives an overview and motivation for our adaptive agent approach, giving context to the remaining sections.  
Section \ref{sec:score} then presents the \colt{} rating function, followed in Section \ref{sec:ace} by the details of the \ace{} method for adapting to teammates. 
Section \ref{sec:ex-eval} experimentally demonstrates the effectiveness of this approach.  
We close with discussion and conclusions in Section \ref{sec:conclusions}. 

\section{Adaptive Approach Overview} \label{sec:overview}
The goal of this work is to design an agent for Codenames that can adapt to a specific teammate.
The approach we adopt is to identify the best language model expert from among a set of options. 
To do this, we frame the problem as a basic \emph{multi-armed bandit} (MAB) problem.
This is an online learning problem where an agent faces, at each time step, a choice among a set of arms, or actions. 
Each arm is assumed to have its own associated distribution over rewards. 
The goal of the agent is to maximize the reward obtained over time. 
At each time step the agent selects an arm and receives a reward drawn from that arm's reward distribution. 
This reward information is utilized by the agent to make arm selection decisions at future time steps. 

For our situation, the set of arms will be the set of Codenames expert agents available within our ensemble agent. 
The popular Upper Confidence Bound (UCB) algorithm \cite{lai1985asymptotically} will be used to determine which expert to select on each turn of the game.
A turn in Codenames consists of the spymaster giving a clue and the guesser making their guesses. 
The expert chosen by UCB will either 1) give a clue (word and number) to our teammate, if acting as the spymaster, or 2) submit guesses in response to the teammate's clue, if acting as the guesser. 
After a turn, the result of the guesses can be observed, providing feedback as to how the chosen expert did. 
This result indicates the number of correct board cards guessed, as well as whether or not a bystander, opponent, or assassin board card was guessed. 
The main challenge in adapting the UCB algorithm to this problem was determining how to incorporate this feedback into a single real-valued reward signal that could be used within the UCB algorithm. 
This is the purpose of the Codenames Linear Team (\colt{}) rating function, which is described in Section \ref{sec:score}.

\subsection{Inspirations}
The \ace{} approach was inspired by several previous ideas, which we now discuss.
Deciding which algorithm to use to solve a specific instance of a problem is a similar challenge that has been addressed previously \cite{leyton2003portfolio}. 
In that research they note that much work in algorithm design is focused on finding a single best algorithm to use in all cases. 
However, for many problems, several good algorithms exist whose effectiveness varies in different portions of the problem space. 
The authors argued that utilizing a portfolio of these good algorithms is a good approach to solving a problem.
Machine learning techniques were used to train a regressor which predicted the run time of each portfolio algorithm on the given problem instance. 
In a similar manner, we feel that using a set of language models has many advantages, as different models will work better for different teammates, and we utilize machine learning to train the \colt{} rating function, which is used to make decisions about which expert to use.

Identifying which from a set of strategies is the best to use against a specific opponent has been addressed in the context of opponent-modeling in poker \cite{bard2013online}. 
In that work the goal was to determine which strategy would get the most reward against the current opponent. 
They approached the problem as a non-stochastic MAB setting, meaning that an adversary chooses the arm rewards.
To address this, the Exp4 algorithm was used \cite{auer2002nonstochastic}, instead of UCB.
Exp4 maintains a probability distribution over a set of experts and assumes that each expert generates its own probability distribution over the possible actions.
The final action at each time step is selected from the weighted mixture of the experts. 
Once the result of that action is observed, an expert's weight is updated based on an estimate of the expected rewards that would have been received if that expert had been chosen during the previous iteration. 
In the case of Codenames our interaction with our teammate is not adversarial, so the non-stochastic MAB setting is not necessary. 
Additionally, the existing AI Codenames agents that will be used as experts each produce a single clue word or sequence of guesses, not a distribution over them, as Exp4 requires. 
For these reasons we use UCB instead of Exp4 in this work, as will be described in more detail in Section \ref{sec:mab}. 

\section{Team Rating Function for Codenames} \label{sec:score}
The design and training of our rating function for evaluating Codenames teams will now be described. 
The goal of having a single real-valued team rating function is motivated by two primary concerns: 1) a single reward function is needed that can be used by UCB within our ensemble framework, and 2) we would like a single dimension to more decisively compare between the performance of different Codenames teams. 

As described in Section \ref{sec:previous-codenames}, prior Codenames research has largely utilized two primary metrics for evaluating the success of teammates in solitaire play: \emph{win rate} and \emph{win time}. 
The \emph{win rate} is the fraction of solitaire games played that are won by the team.  
To win a game the team must correctly flip all of their cards before flipping all of the opponent, bystander, or assassin cards. 
A higher win rate is better. 
The \emph{win time} for the team corresponds to the average number of rounds it takes a team to win a solitaire Codenames game, when they do win. 
A lower win time is better, as it indicates the team is able to win games in fewer rounds.

Two issues arose as we first considered using these metrics as a feedback signal within our adaptive agent. 
\begin{itemize}
    \item Both of those metrics can only be concretely updated after a game has been completed. 
    Without a method to estimate the metrics during a game, there would be no reason to change experts during a game, which means it would take much longer to identify the best expert for a given teammate. 
    We explored some ideas for how to estimate these metrics during a game, which would allow the expert being used to be changed during a game.  
    Other issues like how to decide how much each utilized expert was responsible for the result (win or loss) and the number of turns taken also arose, but ultimately we were stymied by the next major issue with using these two metrics as reward signals. 
    \item Given two teams, with estimates of their corresponding win rates and win time statistics, which is better at the Codenames task? 
    In cases where one team is better than another by both metrics the answer is obvious, but when this isn't the case, it is unclear how to trade-off between these two statistics. 
    What if a team has a slightly lower win rate, but requires fewer rounds on average to win when it does? 
    Would this be better than a team that wins a higher percentage of the time, but takes more turns? 
\end{itemize}
These issues led us to develop our own metric which can rate team performance in Codenames.
Our goal was to find a single numerical metric, correlated with team success in Codenames, which utilizes feedback from individual turns during a game.

\subsection{Features for Codenames}
The proposed rating function takes in a set of features describing the observed performance of a Codenames team and outputs a single number which correlates with how good they are at the Codenames task. 
In this section we address the question of what statistical features will be used as input to the scoring function.

Recalling the previous discussion, these features should ideally reflect things that happen on individual turns in the game of Codenames. 
There are 36 possible outcomes of a single turn in Codenames. 
An outcome of a single turn is completely described by the number of the team's cards that were correctly guessed, which can be any number between 0 and 9, and whether or not an adverse, turn-ending card was guessed. 
The adverse events are guessing an opponent card, guessing a bystander card, both of which end the turn, and guessing an assassin card, which ends both the turn and the game. 
Therefore only one adverse event is possible on a single turn. 
Additionally, if all 9 cards of the team are correctly guessed, then the game is over and the team wins, which means that an adverse event cannot also happen on that same turn. 
So, for each number of team cards guessed (1-8), there are four options (one for each possible adverse event and one for no adverse events)
At least one card must be guessed on each turn, and so for 0 team cards guessed there are three options, one for each adverse event. 
Adding all of these possibilities together gives us 36 total possible outcomes of a single turn. 
All 36 of these outcomes are shown in Table \ref{tab:weights}.

The input to our rating function is a feature vector containing the fraction of the time each of the 36 possible single-turn outcomes occurs for a team when playing Codenames. 
We will denote this probability distribution over outcomes as a vector $X = [x_0, \ldots, x_{35}]$. 
These features can be easily observed as a team plays, during either solitaire or competitive games. 
Each outcome is recorded when it occurs, allowing the probability of each outcome to be estimated. 
This probability estimate will converge to the team's true outcome distribution as more turns are observed. 

\subsection{The \colt{} model}
The proposed \emph{Codenames Linear Team} (\colt{}) rating model is a weighted linear combination of these 36 features that describe a team's performance in Codenames. 

\begin{equation} \label{eq:colt}
\colt{}(X) = \sum_{i \in [0,35]} w_i x_i = W^{\top}X
\end{equation}

The weights $W = [w_0, \ldots, w_{35}]$ of the \colt{} model are trained so that the team ratings produced are meaningful in the competitive game. 
``Meaningful'' is utilized here in the same sense as the Elo ratings, used most notably in Chess \cite{elo1978rating}. 
The Elo rating has an interpretation where the difference in two players' Elo ratings is connected with the probability that one of them wins in a head-to-head matchup. 
The \colt{} scoring function is trained to have the same meaning, so that the difference between the \colt{} ratings for two Codenames teams is predictive of their relative win-percentages in a competitive game between the teams. 

More precisely, let $P_{XY}$ indicate the probability of the team represented by feature $X$ beating the team represented by feature vector $Y$ in a competitive Codenames matchup. 
The trained \colt{} weights should result in
$$
P _{XY} \approx \sigma\left(\colt{}(X) - \colt{}(Y)\right)
$$
where $\sigma(z) = \frac{1}{1+e^{-z}}$.

\begin{figure}[ht]
    \centering
    \includegraphics[width=0.9\textwidth]{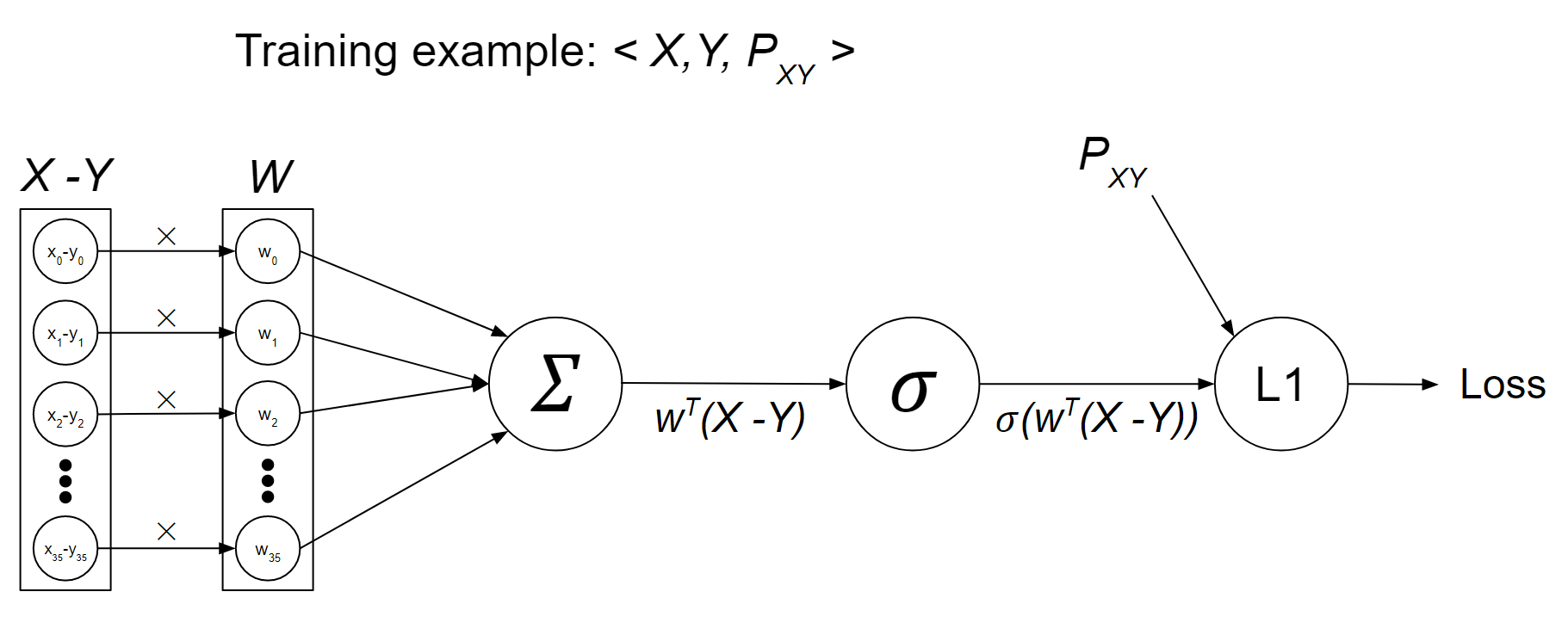}
    \caption{Training the \colt{} model}
    \label{fig:training}
\end{figure}

Since the $\colt{}$ model is a linear combination of features, it is the case that $\colt{}(X) - \colt{}(Y) = \colt{}(X-Y)$.  
The linearity of the \colt{} rating allows us to input the difference of two team's feature vectors and, when the winning percentage between the two teams is known then the target \colt{} rating for that difference vector is also known. 
Having inputs along with the target outputs for the \colt{} function allows the model weights to be trained in a supervised manner. 
Figure \ref{fig:training} shows the structure of the supervised training setup for \colt{}.

To obtain data for training, feature vectors representing probability distributions over turn outcomes were randomly generated\footnote{Full code with details of the \colt{} data generation and training process are available in the supplemental material at \texttt{https://github.com/cjarchibald/codenames-ensemble}}, each corresponding to a possible team. 
Monte Carlo simulations were then used to estimate the winning percentage of each team in a competitive matchup. 
This consisted of simulating a competitive Codenames game by sampling from each team's outcome vector on each turn, according to the given probabilities. 
The Codenames game state was updated with the corresponding randomly sampled outcome, and this process was repeated until the game was over.
1000 simulated games were sampled in this manner for each randomly generated feature matchup in order to produce a winning percentage estimate for the matchup between the two teams. 
The difference in the random feature vectors was used as a training input to the \colt{} model, which then was passed through the $\sigma$ function.
The target output was the estimated winning probability.
18000 training samples were generated in this manner. 
The \colt{} weights $W$ were then trained to convergence using gradient descent with L1 error in PyTorch. 
The final training loss was 0.02699, which led to an $R^2$ score of 0.885.  
To verify that the weights hadn't overfit, the $R^2$ score was calculated on a holdout test set, resulting in 0.883.

The final weights that were obtained are shown in Table \ref{tab:weights}.  
Given these weights $W$ and the feature vector $X$ for a Codenames team, we can calculate the \colt{} rating for the team as $W^{\top}X$.
The \colt{} rating is an integral part of our overall \ace{} approach, which will be described in the next section. 
The \colt{} rating can also be used independently to evaluate Codenames teams and gives a single number which corresponds to a team's ability in the game of Codenames. 

\begin{table}[]
    \centering  
    %Bias term is: 0
\begin{tabular}{|c|c|c|c|c|c|c|c|}
\hline
\textbf{Feature} \hc{}&\hc{} \textbf{Weight} & \textbf{Feature} & \textbf{Weight}&\hc{} \textbf{Feature} &\hc{} \textbf{Weight} & \textbf{Feature} & \textbf{Weight} \\\hline
0100\hc{}&\hc{}-4.695&2010&0.830&4001\hc{}&\hc{}-2.892&7000&2.950\\
\hline
0010\hc{}&\hc{}-1.854&2001&-4.567&5000\hc{}&\hc{}3.022&7100&1.881\\
\hline
0001\hc{}&\hc{}-9.740&3000&2.274&5100\hc{}&\hc{}1.608&7010&2.110\\
\hline
1000\hc{}&\hc{}1.706&3100&0.492&5010\hc{}&\hc{}1.960&7001&-1.806\\
\hline
1100\hc{}&\hc{}-1.637&3010&1.468&5001\hc{}&\hc{}-2.732&8000&2.444\\
\hline
1010\hc{}&\hc{}0.007&3001&-3.798&6000\hc{}&\hc{}2.960&8100&1.120\\
\hline
1001\hc{}&\hc{}-5.551&4000&2.712&6100\hc{}&\hc{}1.792&8010&1.296\\
\hline
2000\hc{}&\hc{}1.941&4100&1.109&6010\hc{}&\hc{}2.129&8001&-1.136\\
\hline
2100\hc{}&\hc{}-0.404&4010&1.945&6001\hc{}&\hc{}-2.573&9000&1.528\\
\hline
\end{tabular}

% \begin{tabular}{|c|c|c|c|c|c|c|c|}
% \hline
% Feature \hc{}& Weight \hc{}& Feature \hc{}& Weight & \hc{}Feature & \hc{}Weight & \hc{}Feature & \hc{}Weight \\\hline
% \hline
% \hc{} 0100&-4.6951&\hc{}2010&0.8299&\hc{}4001&-2.8918&7\hc{}000&2.9496\\
% \hline
% \hc{}0010&-1.8544&\hc{}2001&-4.5665&\hc{}5000&3.0218&\hc{}7100&1.881\\
% \hline
% \hc{}0001&-9.7399&\hc{}3000&2.2741&\hc{}5100&1.6081&\hc{}7010&2.1099\\
% \hline
% \hc{}1000&1.706&\hc{}3100&0.4917&\hc{}5010&1.9599&\hc{}7001&-1.806\\
% \hline
% \hc{}1100&-1.637&\hc{}3010&1.4676&\hc{}5001&-2.7322&\hc{}8000&2.4437\\
% \hline
% \hc{}1010&0.0066&\hc{}3001&-3.7975&\hc{}6000&2.96&\hc{}8100&1.1205\\
% \hline
% \hc{}1001&-5.5511&\hc{}4000&2.7124&\hc{}6100&1.7916&\hc{}8010&1.2955\\
% \hline
% \hc{}2000&1.9413&\hc{}4100&1.1092&\hc{}6010&2.1286&\hc{}8001&-1.1355\\
% \hline
% \hc{}2100&-0.4042&\hc{}4010&1.9446&\hc{}6001&-2.5731&\hc{}9000&1.528\\
% \hline
% \end{tabular}

     \caption{Learned feature weights for \colt{}.  Each feature describes the outcome by the number of words flipped of type \emph{team}, \emph{opponent}, \emph{bystander}, and \emph{assassin}. For example, feature 2010 indicates the event where two \emph{team} cards are flipped, followed by one \emph{bystander} card.}
    \label{tab:weights}
\end{table}

We note that the specific \colt{} weights shown in table \ref{tab:weights} would likely change if trained on a different dataset, perhaps specific to a population, rather than synthetically generated as was done here. 
Some of the outcomes, like guessing 9 correct cards on a single turn, are generally very unlikely to occur and as such, there isn't a lot of data that can be used to influence those weights. 
This leads to that weight being smaller than others even though in theory it is the best possible outcome.
While some of the specific weights might change when trained with another data source, the result of following this same training methodology should still result in a meaningful and useful \colt{} rating. 
In the remainder of this paper the \colt{} rating function will refer specifically to \colt{} using the weights in Table \ref{tab:weights}.

\subsection{\colt{} Exploration}
\begin{figure}[h]
    \centering
    \includegraphics[width=0.8\textwidth]{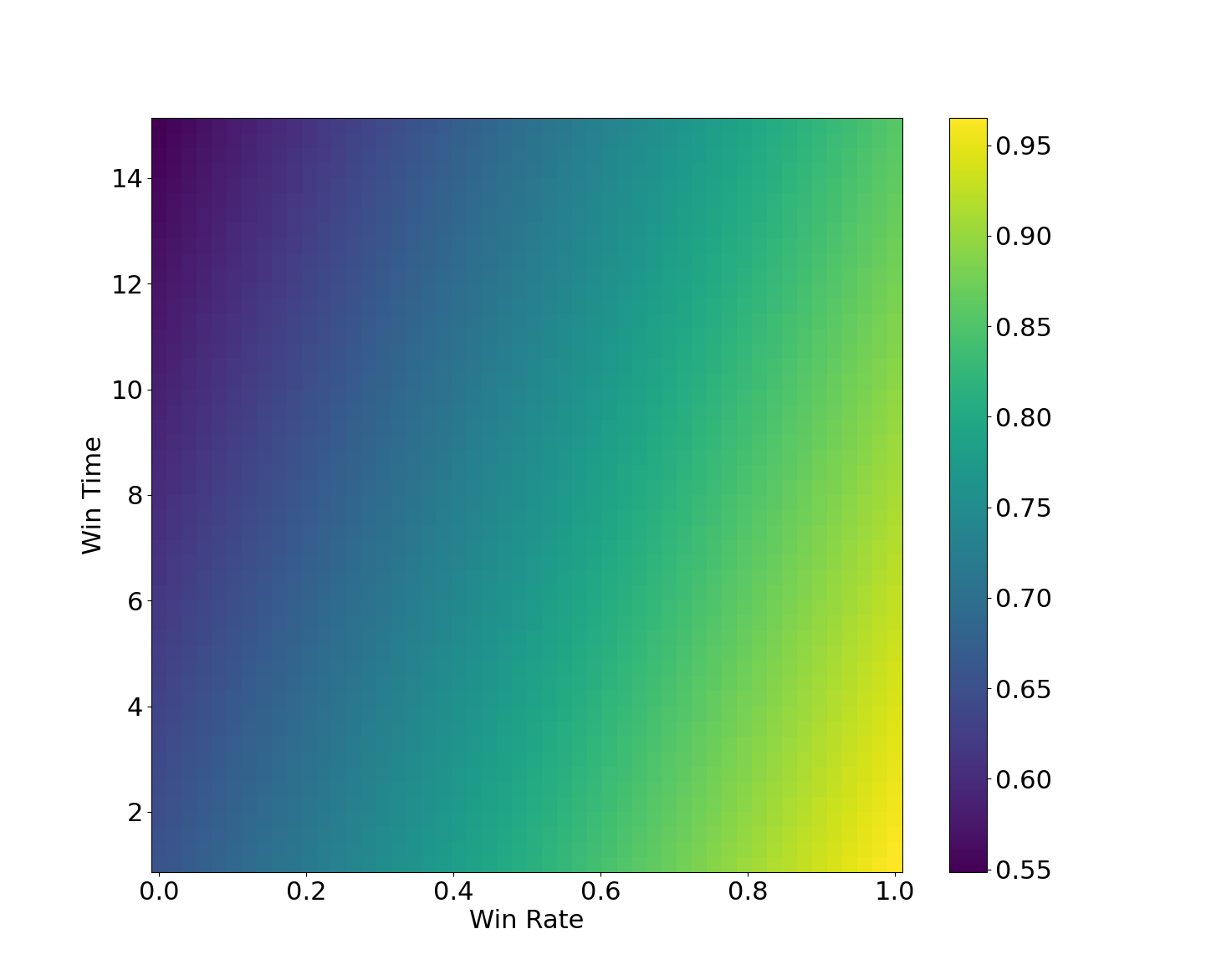}
    \caption{Estimated average \colt{} rating as a function of average win rate and win time}
    \label{fig:colt-surf}
\end{figure}

As one purpose of the \colt{} rating function is to be used in place of the two previous metrics, in this section the relationship between \colt{} rating, win rate, and win time will be explored. 
To do this, Monte Carlo simulation, similar to that used to generate the training data for \colt{}, was used. 
Team outcome distribution vectors were randomly generated, and 1000 solitaire games were simulated, sampling turn outcomes from the given distribution vector. 
The \colt{} rating was generated for each of these vectors, along with the average win rate and win time estimated from the 1000 simulated games. 
This process was repeated for 500 different feature vectors. 
A radial basis function interpolator was used with a Gaussian kernel to estimate the average \colt{} rating as a function of win rate and win time. 
The estimated average \colt{} rating is plotted as a function of win rate and win time in Figure \ref{fig:colt-surf}. 

It is clear that both a higher win rate and a lower win time lead to a higher \colt{} rating, as expected.
The way that the \colt{} rating changes gives a sense of how the two metrics interact with each other and can be used to compare previously incomparable pairs of statistics about a Codenames agent. 
As an example, from this figure we can find many different combinations of win rate and win time that yield approximately the same average \colt{} rating. 

A \colt{} rating of 0.86 coincides with an average win rate of 0.94 and an average win time of 11.75 turns, or a win rate of 0.80 with 6.25 turns, or a win rate of 0.70 with 2.60 turns.  
A \colt{} rating of 0.90 can be achieved with an average win rate, win time combo of (0.99, 8.41), (0.88, 3.93), or (0.85, 2.62). 
In both cases as the win rate decreases the win time must also decrease in order to maintain the same overall \colt{} score. 
Without the \colt{} score, it wouldn't be clear how much decrease in win time is required to offset a decrease in win rate of a given amount, but now this can be done with the \colt{} score. 
It gives us a single dimension to compare different Codenames team performance. 
It can be used on teams formed with only AI agents or teams including human agents as well. 
The only requirement is that all of a team's turn outcomes are recorded.

In particular, we feel that the \colt{} score could be used as a single number which captures the compatibility of two language models. 
If we play many Codenames games between two AI agents, each using one of the language models, the \colt{} score produced from their play outcome distribution captures how well they do at Codenames, which in turn depends on how compatible the language models are. 

\section{An Adaptive Ensemble Agent for Codenames} \label{sec:ace}
Given the novel \colt{} rating, the proposed Adaptive Codenames Ensemble (\ace{}) method can now be described. 
\ace{} utilizes a set of Codenames agents, which will be referred to as \emph{experts}, each of which can do everything required for the game. 
The set of expert agents can use different language models and/or strategies for generating clues and guesses in Codenames. 
Our \ace{} agent will be matched with an unknown teammate and the goal is to maximize performance by determining which expert to use on each turn. 
The subsequent outcome of each turn will be observed, the \ace{} agent will update its internal statistics about the experts, and the entire process will be repeated.

\subsection{Multi-Armed Bandits} \label{sec:mab}
We frame the task facing the \ace{} as a \emph{multi-armed bandit} (MAB) problem.  
In a traditional multi-armed bandit problem, an agent faces a choice of arms, or actions, to select at each time step. 
When an arm is selected, a random reward is drawn from the corresponding reward distribution. 
The goal of a MAB algorithm is to maximize cumulative reward over time, or equivalently, to minimize the regret the agent has for not choosing the best arm at each time step. 

Many different problems and tasks have been modeled as MABs. 
As a result, there are many MAB algorithms designed to determine which arm to pull at each time step.
A survey of applications of the MAB model can be found in \cite{bouneffouf2020survey}, while an overview of MAB algorithms can be found in \cite{lattimore2020bandit}.  
In this work we utilize one of the more popular algorithms, the Upper Confidence Bound (UCB) algorithm \cite{lai1985asymptotically}, which was notably used within the Monte Carlo Tree Search framework \cite{kocsis2006bandit}.
The UCB algorithm has an expected cumulative regret that is logarithmic in the number of time steps. 
UCB maintains for each arm $i$ a record of how many times that arm has been pulled ($n_i$) and also the total reward that has been obtained when that arm was pulled ($r_i$). 
If we let $N = \sum_i n_i$, then at each time step, the UCB algorithm selects the arm that maximizes: 
\begin{equation}\label{eq:ucb}    
UCB_i = \frac{r_i}{n_i} + c \sqrt{\frac{\log N}{n_i}}
\end{equation}
$c$ is a parameter of the UCB algorithm, and typically needs to be tuned for a specific application and its rewards. 

\subsection{The Adaptive Codenames Ensemble (\ace{}) Algorithm}
The \ace{} algorithm shown in Algorithm \ref{alg:ace} treats each expert as an arm and uses the UCB algorithm to determine which expert to select on each turn. 
A vector of counts ($C_i$) is maintained for each expert which indicates how many times each turn outcome was experienced when that expert agent was selected. 
$n_i$ for each expert can be calculated as the sum of those counts. 
The overall number of turns experienced is also maintained as $N$. 
At each time step, the count vector $C_i$ is normalized into a probability distribution and passed into the \colt{} rating function, producing a ratings estimate $r_i$ for each expert in the ensemble. 
This is utilized as the $r_i$ in the UCB algorithm shown in Equation \ref{eq:ucb}. 
At each time step the expert with the highest UCB score is selected, and its clue or guesses are utilized in the game. 
When the outcome of that turn is observed, the relevant counts for that agent are updated. 
Counts for other experts that generated the exact same clue or guesses can also be updated. 

\begin{algorithm}
\caption{The Adaptive Codenames Ensemble (\ace{}) algorithm}\label{alg:ace}
\begin{algorithmic}
\Require set of $m$ experts $E = \{1, 2, \ldots m\}$
\State $\mathbf{C_i} \gets \left [0, \ldots, 0 \right ]$ for each $i \in E$ \Comment{Length 36 zero vector}

\State $N \gets 0$
\While{Game continues}
\ForAll{$i \in E$}
\If{$n_i == \infty$}
\State $UCB_i \gets \infty$
\Else
\State $UCB_i \gets \colt{}\left(\frac{\mathbf{C}_i}{n_i}\right) + c \sqrt{\frac{\log N}{n_i}}$ 
\LComment{$\frac{\mathbf{C}_i}{n_i}$ indicates element-wise division of vector $\mathbf{C}_i$ by $n_i$} 
\EndIf
\EndFor
\State $q \gets \argmax_i UCB_i$ \Comment{Break ties randomly}
\State Play expert $q$ for this turn, observe outcome $j$
\State $\mathbf{C}_q[j] \gets \mathbf{C}_q[j] + 1$ 
\LComment{Can also update counts for all experts with exact same clue or guesses}
\State $N \gets  N + 1$
\EndWhile
\end{algorithmic}
\end{algorithm}

\subsection{\ace{} Discussion}

We want to briefly comment on one difference between the \ace{} algorithm as presented and the typical way that UCB is utilized. 
The UCB calculation in algorithm \ref{alg:ace} is written in such a way that the \colt{} rating is computed each time on the new normalized count vector.  
UCB is typically described as storing the cumulative rewards and selection count for each arm and then computing the average reward from those counts. 
The \ace{} algorithm could be modified to instead give a reward for selecting agent $k$ of $w_j$ after observing outcome $j$ when agent $k$ was used. 
The computed UCB values for the arms would be unchanged in this alternate formulation.

\section{Experimental Evaluation} \label{sec:ex-eval}

To experimentally evaluate the performance of the proposed \ace{} agent several experiments were conducted. 
In each case an \ace{} agent, internally utilizing a group of experts, was partnered with another agent, utilizing a single language model or approach. 
The experiments differ in which agents are present in the group of experts when facing a given partner. 
Experiments were performed with the \ace{} agent as the spymaster and also as the guesser. 

\subsection{Experimental Setup} \label{sec:ex-setup}
Each experiment had a specific pair of agents, spymaster and guesser, forming a team. 
This team would play 50 consecutive games and the statistics of the team performance during these 50 games were tracked. 
This process was repeated for the adaptive \ace{} agent teams approximately 580 times, while the teams composed of only static base agents completed 1500 sets of 50 games, for 85000 games per pair. 
The focus in our results is on the average performance of each team across these repetitions. 
Each team faced the same sequence of game boards to ensure fair comparison.
The only parameter of the \ace{} algorithm is the UCB constant $c$. 
This constant was set to 0.5, based on a parameter sweep using data from simulations of games that were generated separately from the experimental results detailed next.  

The \ace{} agent was evaluated in two main different cases. 
\begin{enumerate}
    \item \textbf{With Partner}: in this case the matching partner for their current teammate is in the ensemble.  
    By matching partner, we mean an agent utilizing the same language model as the teammate, as will be described in Section \ref{sec:agents}. 
    For the \emph{with partner} experiments, the \ace{} agent always has the same set of agents in its ensemble, including the agent with the same language model as any partners it would face.
    Since the performance of an agent with its matching partner is often significantly better than with any other agent, the goal of these experiments is to see how well the \ace{} agent is able perform when there is a clear best expert option. 
    \item \textbf{Without Partner}: the matching partner is not in the ensemble.  
    This is the case where the ensemble does not contain the matching partner for the current teammate. 
    This could be due to the fact that it is an unknown AI agent, or the partner could be a human agent.  
    This could occur in ad hoc team settings where when the ensemble is created the exact future teammates and their language models are unknown. 
    For the \emph{without partner} experiments, the \ace{} agent will exclude from the set of experts the matching partner for the current teammate. 
    Since the performance of the agent teams can be a lot more varied without matching partners, the goal in this set of experiments is to determine how well the \ace{} agent performs when it might be harder to differentiate between the performance of the various agents in the ensemble.
\end{enumerate}

\subsection{The Base Codenames Agents} \label{sec:agents}
The same set of base Codenames agents are used in the experiments both as experts within the \ace{} agent and as teammates. 
All of these base agents utilize the same strategy to determine what clues to give as spymaster and which cards to guess when the guesser. 
The agents differ, however, in the internal language model used. 
In this section the common strategy of these agents will be briefly described\footnote{Full code and details available at \texttt{https://github.com/cjarchibald/codenames-ensemble}}, followed by details about the language models used. 

\subsubsection{Base Agent Strategy}
The base agents utilize strategies that are very similar to what was described in the original Codenames paper \cite{kim2019cooperation}, 
At a high level, the strategy for the spymaster is to find a clue word that is close to a subset of its team's board words and far from all other words on the board. 
The guesser, when given $k$ words to guess, guesses the $k$ nearest board words to the clue it is given. 
Cosine distance is used to determine distance between different words. 

The spymaster strategy we use was modified from that described in previous work. 
This modification was done to give the agent greater flexibility regarding the number of words associated with a clue and the number of guesses made. 
Prior work did this either by fixing the maximum clue number or by setting a distance threshold, which had largely the same result of specifying a maximum clue number. 
We didn't want to limit our agents to an arbitrary maximum clue number and also found that distance thresholds worked inconsistently across different language models due to the different dimensionality of the language models, which caused the cosine distances between words to be on different scales for the different models. 

Our modified approach was not to use an explicit threshold, but rather to determine the best clue number based on the current board words and how close other board words are to the teams board words and potential clues. 
One factor preventing this from being done in previous work was the computation cost of searching all possible clues (all words in the corpus) and their distances to the current board words. 
This was addressed in our work by precomputing the 300 nearest neighbors in every language model for the subset of 10575 English words that were common among all the language models. 
We chose to use 300 associations for each word based on preliminary tests focused on optimizing the performance of the base agent teams. 

The spymaster selects a clue by considering these 300 most associated words for each of their team's board words (the good words).
Each associated word is treated as a potential clue. 
The spymaster keeps track of every potential clue and its distance to the team's board words. 
For each possible clue the spymaster also determines the distance to the closest bad word on the board (word not belonging to the team). 
The spymaster then filters the list of associated board words for each potential clue by removing any good board words that are farther away than the closest bad word. 
The spymaster then selects the clue with the highest number of associated good board words. 
In the case of a tie, the spymaster chooses the potential clue with the smallest average distance from each remaining associated good board word to the potential clue.
Since the guesser simply guesses the closest $k$ board words to the clue word, when using the same language model as the spymaster there will be no incorrect guesses.

This spymaster approach that uses associated word lists allows the base agents to exhibit flexibility in clue number.
When bad words are close to clues for the good words it can give a clue for only 1 word.
When many good words are close to the same clue it can take advantage and give clues for many more board words, even up to 9, if the board permits. 

We compared the performance of these basic agents with the original agents described in \cite{kim2019cooperation}. 
In that work they used 3 separate thresholds to control the `aggressiveness' of the agent: low, medium, and high.
The low threshold had the highest win rate and highest win time, while the high threshold had a much lower win rate but also lower win time.
They reported win rates and win times for different agent pairs across 30 games, which still leaves high variance in the results. 
For our base agents, using the associated word lists, we evaluated teammate pairs using 85000 games.
The main purpose of this comparison was to ensure that the general performance of our agents was consistent with that reported in previous work.
While acknowledging the fact that a statistically significant comparison is not possible, given the vastly different number of experimental games used, we made a few general observations\footnote{Full comparison details available at \texttt{https://github.com/cjarchibald/codenames-ensemble}}.
First, our base agents generally had win rates that were lower than their medium agents, but higher than their high agents. 
Second, our base agents had win times that were lower than their most aggressive high threshold agents. 
Taken together this shows that our approach was most similar to their high threshold agents, but seemed to constitute an improvement in both dimensions. 
It also has the advantages of not requiring a threshold parameter to be specified and of running faster. 

\subsubsection{Base Agent Language Models}
The basic agents included in our experiments all used this same strategic framework, differing only in the word embedding used to compute distances between words. 
Each language model specifies a high-dimensional embedding for each word that can appear on the board or as a clue. 
This embedding is used to determine distances between the different words and these distances are used by the strategy to make decisions about what actions to select. 
Each base agent word embedding is listed below, along with the letter or letters that are used to designate it in the experimental results tables and figures.  

\begin{itemize}
    \item word2vec (w) - is a 300 dimensional word embedding described in \cite{mikolov2013distributed} which was used previously for Codenames in \cite{kim2019cooperation}.
    \item GloVe (g$k$) - was introduced in \cite{pennington2014glove} and first used for Codenames in \cite{kim2019cooperation}. It can generate different dimensionality embeddings. To represent closely related language models, 4 different dimensionality GloVe agents were included.  They are GloVe-50 (g5), GloVe-100 (g1), GloVe-200 (g2) , and GloVe-300 (g3), where in each case GloVe-$k$ refers to a $k$-dimensional GloVe embedding.
    \item word2vec+GloVe (wg) - is a concatenation of word2vec and GloVe embeddings, which was indentified as a successful approach in \cite{kim2019cooperation}.   
    To include agents with overlapping, but not identical, language models in our experiments, we use a word2vec+GloVe-300 language model as the guesser, while for the spymaster word2vec was concatenated with GloVe-50.  
    These two embeddings overlap for part of the dimensionality, but differ elsewhere.
    As the guesser and spymaster embeddings aren't exactly the same, they were not treated as matching partners in the experiments. 
    \item ConceptNet NumberBatch (cn) - is an embedding that is derived from both ConceptNet and distributional embeddings learned from text.
    ConceptNet is a knowledge graph language representation that connects words to related words via concepts. 
    Both the ConceptNet and ConceptNet Numberbatch language models were presented in \cite{speer2017conceptnet}.
    To our knowledge, this is the first time that ConceptNet Numberbatch has been applied to Codenames. 
\end{itemize}

These agents were chosen with the intent of being a representative, but not exhaustive, sample of approaches from previous Codenames research. 
All of the language models used, with the exception of the ConceptNet NumberBatch embedding, can be found in multiple previous works. 
Some of the language models are closely related, while others are wildly different. 
As will be seen in the results, some are fairly compatible, while others work terribly as teammates.
The experiments will show how the proposed methods work with this group of agents, but there is no reason it wouldn't work with other expert agents, since the \colt{} and \ace{} methods do not depend at all upon how the experts make their decisions. 
Any existing agent, or future novel agents, could be added to the expert ensemble and the resulting \ace{} agent should be effective.

\subsection{Experimental Evaluation} \label{sec:evaluation}
How can we tell from the experimental results how well the \ace{} agent is working?
What are reasonable things that it can be compared to? 
As \ace{} is the first adaptive agent proposed for Codenames, a previous, similarly adaptive agent does not exist for comparison. 
The upper bound on performance for the \ace{} agent is the performance of the best expert in its ensemble with an individual partner. 
Ideally, the performance of the \ace{} agent should be close to that of the best expert. 
As will be shown, how close the \ace{} agent can get to that upper bound performance will depend on the composition of the ensemble as well as the performance of the other ensemble experts with the specific teammate. 
If most experts perform very poorly with a teammate, the ensemble score will be negatively impacted during the time when it is exploring the experts and giving them chances to play with the teammate. 
In our experimental results, in addition to this upper bound, we will compare \ace{} to two other approaches. 

The first, which will be called the \emph{best average} (BA) agent, will approach the problem from the following standpoint. 
Assume that the set of possible agent teammates was known, and that the set of available experts was also known and accessible. 
One reasonable approach would be to evaluate every expert agent with every possible teammate and then choose the expert that has the best average performance with teammates, given some probability of being paired with each teammate. 
This best average expert could be used to play with every new teammate, since the identity of teammates when playing would be unknown.  
This agent would perform the best in the long run, compared to other single expert agents that could be selected. 
In the experimental results, we will assume that the set of experts among which we are choosing the best average agent is the same as the set of experts available to the \ace{} agent. 
Additionally we will assume a uniform distribution over the possible teammates we could be paired up with. 

The downsides and complications of this approach are that it assumes that the set of possible teammates is known and that the probability of being paired with each teammate is known. 
Beyond this knowledge, it is also assumed that the performance of every possible expert with every potential teammate is known or can be precomputed. 
These are obviously strong assumptions, but ones that are satisfied in our experimental setup. 
In contrast, the \ace{} agent makes no assumptions about the set of teammates and utilizes no prior knowledge about how different experts might perform with various teammates. 
It learns all of this information as it plays and can adapt which expert is chosen as it gains experience with a given teammate. 

In some cases the best average agent will in fact be the best possible expert for a given teammate, or very close to it, and in these cases we don't expect the \ace{} agent to outperform the best average agent. 
Rather, we anticipate that when performance across all possible teammates is combined, the \ace{} agent will do better on average, and that it is thus a simpler choice which yields superior results. 

The other approach to which the \ace{} agent will be compared is a \emph{random agent} (R).  
This agent will have the same set of expert agents available as does the \ace{} agent. 
The difference is that, instead of utilizing the \colt{} rating to make an informed decision about which expert to choose on each turn, the random agent will simply choose among the experts uniformly at random.  
This gives in some sense a baseline for the performance of the ensemble, if it never learned anything about the experts. 

\subsection{Experimental Results} \label{sec:ex-results}

The results of the experiments which were described in Section \ref{sec:ex-setup} are now presented and discussed.
The main results are shown in Tables \ref{tab:coltx}, \ref{tab:wrx}, and \ref{tab:wtx}.
Each of these tables has the same format, but each reports the results according to a different metric, the \colt{} rating, win rate, and win time, respectively.
These results are the average performance of the various agent teams across all 50 games in each experiment, averaged across all experiments that were performed. 
The largest 95\% confidence interval across all of these values is shown in the caption for each table.
Any difference of more than twice this confidence interval in the tables can be considered to be statistically significant. 

Each table has spymasters (SM) as rows and guessers as columns. 
The entry in each cell indicates the performance of the team formed by the spymaster and guesser corresponding to that row and column of the table, according to the metric of the table. 
These results for the basic agents form the upper left portion of the table. 
The text within a cell is bolded if it is the best value of that metric for either the spymaster or guesser agent, in either the \emph{with partner} or \emph{without partner} case. 

A look at the results between these basic agents shed light on the composition of this specific agent population and some of the challenges the adaptive agent faces. 
First, each agent, with the exception of word2vec+GloVe (wg), which has slightly different language models for the spymaster and guesser, all agents perform the best by every metric when paired with a teammate using the same language model. 
Second, it is clear that, generally speaking, the GloVe-based agents are fairly compatibly with each other, performing decently in most cases. 
Third, the word2vec (w)  and ConceptNet Numberbatch (cn) agents generally are not very compatible with any other agents. 
Interestingly, the highest \colt{} score in Table \ref{tab:coltx} is when the cn spymaster is paired with the cn guesser, but with every other agent as guesser, it has a negative \colt{} rating. 
All of the win rates for the cn spymaster with other guessers are between 59-77\%, which is not very good. 
The \ace{} agent will have some subset of these base agents as its experts to choose from, and so in some cases will have several good options, while in other cases there might only be one or no good options. 

The results for the \ace{}, BA, and R guessers are shown in the upper right, for both the \textit{with} and \textit{without partner} cases. 
The bottom left of the table shows the results for the \ace{}, BA, and R spymasters, again in each partner case.
Each of these sections of the table has the same set of five divisions. 
The first division, labeled `A', indicates the average performance of the spymaster or guesser across all of the teammates within that category.  
The second division, labeled `B' and colored blue, indicates the best performance of that agent across all teammates within the given category (with or without partner). 
This indicates the best possible performance that the \ace{} agent could aspire to. 
The next division shows the performance of the \ace{} agent with the given teammate. 
The `BA' division shows the performance of the best average agent, which was described in section \ref{sec:evaluation}.
This would be the best that could be done if a single agent had to be picked to work with all the teammates.  
The final division is `R', which indicates the random agent, also described previously.
Each section of the table additionally has 4 cells that jut out, and these cells show, in bold, the performance of the best agent (B), \ace{}, the best average agent (BA), and the random agent (R), averaged across all of the teammates. 
The best performer out of \ace{}, BA, and R, is highlighted in brighter yellow. 
This average across all teammates provides, for each case, a single statistic that summarizes performance against the teammate population used in these experiments. 

\subsubsection{With Partner}
We will first examine the results when the \ace{} agent has all 7 available experts in the ensemble. 
In particular, this means that when paired with a teammate, the matching partner, using the same language model, is one of the experts that can be chosen. 
The lone exception is the word2vec+glove agent (wg), for which, as previously discussed, the language models used are slightly different for the guesser and spymaster. 

First, how does the \ace{} agent do as the guesser? 
This can be seen in the top right part of the table, under the `With Partner' heading. 
Each row of the table corresponds to a particular spymaster (SM).
The blue `B' column shows the best performing static guesser with that spymaster. 
The next three columns (`\ace{}', `BA', and `R') show the performance of the \ace{} agent, best average guesser, and random ensemble agent with this spymaster. 
For the \colt{} rating shown in table \ref{tab:coltx} the best guesser on average, across all the static spymasters, is the wg agent, which can be seen by the `A' row underneath its column being the highest (0.80 in the case of the \colt{} rating shown in table \ref{tab:coltx}). 
A yellow highlight indicates which of these three approaches performs best with that row's spymaster.
According to all three metrics, \ace{} is the best overall guesser for every single spymaster agent, better than both BA and R. 
\ace{} is often better by a wide margin, and gets very close to the performance of the best agent from the ensemble. 
The win rates in Table \ref{tab:wrx}, the results are particularly close.

As the spymaster, we can see that again \ace{} always does better than the best average and random agents by \colt{} rating.
The one exception is the case when the best average agent is paired up with its perfect partner, but even then, \ace{} gets very close. 
Looking at the average performance across all teammates, we can see that \ace{} does much better than the best average agent (BA) in the \emph{with partner} case. 
When teammates using the same language model are paired together, the performance is usually significantly better than when non-matching teammates are paired. 
In the \emph{with partner} experiment cases, this gives a very strong signal to the \ace{} agent, via the \colt{} score, that enables the best partner to be identified quickly and clearly, leading to very good performance from the \ace{} agent. 

The results in Tables \ref{tab:wrx} and \ref{tab:wtx} generally tell the same story in the \emph{with partner} case: close to optimal performance by \ace{} both in terms of win rate and win time, when acting as both the spymaster and guesser. 
These results additionally confirm that explicitly maximizing the \colt{} score leads to good performance according to both of the other metrics as well. 

\subsubsection{Without Partner}
In the \emph{without partner} case we are excluding the matching partner agent from the ensemble. 
This means that there is very often not an expert in the ensemble that stands out from the others in how it performs with the current teammate. 
In fact, all of the expert agents might perform very poorly with a given teammate.
We expect this to generally make things harder on the \ace{} agent to get close to the best possible performance. 
For these reasons we again evaluate the \ace{} agent based on how close it is able to get to the performance of the best partner in its ensemble. 
In particular we compare to the best average teammate agent, excluding the matching partner from consideration. 
Each agent's average performance is evaluated across all agent's except their matching partner.

We first focus on the \colt{} score results, shown in Table \ref{tab:coltx}.
As the guesser, we see that when paired with 1 of the 7 spymasters, the \ace{} agent performs better than the best average guesser and the random ensemble guesser. 
In this cases it does quite a bit better than the best average static agent, which in this case is the word2vec+GloVe (wg) agent.
For the other 6 cases, the best average agent does better than the \ace{} agent. 
However, for all of these the \ace{} score is relatively close. 
When playing with the wg guesser, \ace{} outperforms the best average agent by 0.47. 
When the best average is better, the biggest difference is 0.22, in the case of the g2 guesser.

The bold yellow cells underneath the `Without Partner' `\ace{}' and `BA' columns show the performance of each of these guesser approaches, averaged across all of the spymasters in the population. 
The best average agent achieves an average \colt{} rating of 0.80, which is slightly higher than the average rating obtained by the \ace{} guesser.
In Table \ref{tab:wrx} we can see that the average win rate across all spymasters is the same for each approach. 
Table \ref{tab:wtx} shows that the best average approach obtains a slightly lower average win time than \ace{} across this population. 

As the spymaster, the \ace{} agent performance is similar, but better in comparison to the best average agent.
We starting again with the \colt{} results shown in Table \ref{tab:coltx}.
In 5 of the 7 cases the \ace{} spymaster outperforms the best average agent, by an average amount of 0.39. 
In the other 2 cases, where the best average agent is better, it's average difference is 0.20.
This leads to the \ace{} spymaster's average \colt{} rating of 0.86 being much higher than the rating of 0.63 for the best average agent. 
Again, looking at win rate and win time in tables \ref{tab:wrx} and \ref{tab:wtx} show similar results, with \ace{} doing the best overall. 

Taken all together, these results show that the \ace{} agent is able to achieve nearly the same performance as the best average agent in the case where it is the guesser, while outperforming it when acting as the spymaster. 
Importantly, this performance comes with far fewer assumptions. 

In order to accurately select the best average agent for a given population we must know exactly which teammates we might face and how well each expert in the ensemble will do against each possible teammate. 
The \ace{} agent, on the other hand, requires no knowledge of possible teammates or prior information as to how well each expert will do. 
It simply adjusts its play based on feedback received during play with a given teammate. 
All that is required is to determine the set of experts and start playing.  
Thus, the \ace{} agent is a much easier way to get good performance across a set of teammates. 

\begingroup
\setlength{\tabcolsep}{5pt}
\begin{landscape}
\vspace*{\fill}
\begin{table}[ht]
\begin{center}
    \begin{tabular}{|c|c|c|c|c|c|c|c|c|c|c|c|c|c|c|c|c|c|}
\hline
& \multicolumn{7}{c|}{\textbf{Guessers}} & \multicolumn{5}{c|}{\textbf{With Partner}} & \multicolumn{5}{c|}{\textbf{Without Partner}}\\
 \hline
\textbf{SM}& \textbf{w} \hc{}& \textbf{wg} \hc{}& \textbf{g5} \hc{}& \textbf{g1} \hc{}& \textbf{g2} \hc{}& \textbf{g3} \hc{}& \textbf{cn} \hc{}& \textbf{A} \hc{}& \textbf{B} \hc{}& \textbf{ACE} \hc{}& \textbf{BA} \hc{}& \textbf{R} \hc{}& \textbf{A} \hc{}& \textbf{B} \hc{}& \textbf{ACE} \hc{}& \textbf{BA} \hc{}& \textbf{R} \hc{}\\ \hline 
\textbf{w} \hc{}& \textbf{2.08}& 0.23& -0.93& -0.66& -0.42& -0.33& \textbf{0.36}& 0.05& 2.08\bc{}& 2.04\bw{}& 0.23& 0.19& -0.29& 0.36\bc{}& 0.20& 0.23\bwo{}& -0.19\\ 
 \hline 
\textbf{wg} \hc{}& \textbf{-0.04}& 0.85& \textbf{1.60}& 0.85& 0.62& 0.66& -0.30& 0.61& 1.60\bc{}& 1.32\bw{}& 0.85& 0.69& 0.61& 1.60\bc{}& 1.32\bwo{}& 0.85& 0.69\\ 
 \hline 
\textbf{g5} \hc{}& -0.81& 0.32& \textbf{2.05}& \textbf{0.61}& 0.26& 0.26& -0.93& 0.25& 2.05\bc{}& 2.00\bw{}& 0.32& 0.36& -0.05& 0.61\bc{}& 0.30& 0.32\bwo{}& 0.04\\ 
 \hline 
\textbf{g1} \hc{}& -0.46& \textbf{0.97}& 0.87& \textbf{2.05}& 0.96& 0.95& -0.36& 0.71& 2.05\bc{}& 1.98\bw{}& 0.97& 0.81& 0.49& 0.97\bc{}& 0.81& 0.97\bwo{}& 0.58\\ 
 \hline 
\textbf{g2} \hc{}& -0.34& 1.49& 0.41& \textbf{0.88}& \textbf{2.05}& \textbf{1.51}& -0.14& 0.84& 2.05\bc{}& 1.98\bw{}& 1.49& 0.94& 0.64& 1.51\bc{}& 1.27& 1.49\bwo{}& 0.74\\ 
 \hline 
\textbf{g3} \hc{}& -0.36& \textbf{1.84}& 0.17& 0.64& \textbf{1.26}& \textbf{2.09}& -0.05& 0.80& 2.09\bc{}& 2.00\bw{}& 1.84& 0.89& 0.58& 1.84\bc{}& 1.66& 1.84\bwo{}& 0.69\\ 
 \hline 
\textbf{cn} \hc{}& -0.15& \textbf{-0.10}& -1.15& -0.82& -0.48& -0.33& \textbf{2.12}& -0.13& 2.12\bc{}& 2.05\bw{}& -0.10& -0.02& -0.51& -0.10\bc{}& -0.22& -0.10\bwo{}& -0.44\\ 
 \hline 
\multicolumn{8}{c}{\textbf{With Partner}} &&\textbf{2.01}\bcf{}&\textbf{1.91}\bwf{}&\textbf{0.80}&\textbf{0.55}&&\textbf{0.97}\bcf{}&\textbf{0.76}&\textbf{0.80}\bwf{}&\textbf{0.30}\\ \cline{1-8}\cline{10-13}\cline{15-18}
\textbf{A}\hc{}& -0.01& 0.80& 0.43& 0.51& 0.61& 0.69& 0.10& \multicolumn{9}{c}{} \\ \cline{1-9}
\textbf{B}\hc{}& 2.08\bc{}& 1.84\bc{}& 2.05\bc{}& 2.05\bc{}& 2.05\bc{}& 2.09\bc{}& 2.12\bc{}&\textbf{2.04}\bcf{}& \multicolumn{9}{c}{} \\ \cline{1-9}
\textbf{ACE}\hc{}& 2.00\bw{}& 1.64\bw{}& 1.96\bw{}& 1.96\bw{}& 1.96& 2.00\bw{}& 2.03\bw{}&\textbf{1.94}\bwf{}& \multicolumn{9}{c}{} \\ \cline{1-9}
\textbf{BA}\hc{}& -0.34& 1.49& 0.41& 0.88& 2.05 \bw{}& 1.51& -0.14&\textbf{0.84}& \multicolumn{9}{c}{} \\ \cline{1-9}
\textbf{R}\hc{}& 0.14& 0.83& 0.54& 0.57& 0.67& 0.76& 0.20&\textbf{0.53}& \multicolumn{9}{c}{} \\ \cline{1-9}
\multicolumn{8}{c}{\textbf{Without Partner}} & \multicolumn{10}{c}{} \\ \cline{1-8}
\textbf{A} \hc{}& -0.36& 0.80& 0.16& 0.25& 0.37& 0.45& -0.24& \multicolumn{10}{c}{} \\ \cline{1-8}
\textbf{B} \hc{}& -0.04\bc{}& 1.84\bc{}& 1.60\bc{}& 0.88\bc{}& 1.26\bc{}& 1.51\bc{}& 0.36\bc{}&\textbf{1.06}\bcf{}& \multicolumn{9}{c}{} \\ \cline{1-9}
\textbf{ACE} \hc{}& -0.16\bwo{}& 1.65\bwo{}& 1.41\bwo{}& 0.73& 1.00\bwo{}& 1.26& 0.10\bwo{}&\textbf{0.86}\bwf{}& \multicolumn{9}{c}{} \\ \cline{1-9}
\textbf{BA} \hc{}& -0.34& 1.49& 0.41& 0.88\bwo{}& 0.62& 1.51\bwo{}& -0.14&\textbf{0.63}& \multicolumn{9}{c}{} \\ \cline{1-9}
\textbf{R} \hc{}& -0.22& 0.83& 0.25& 0.32& 0.44& 0.53& -0.14&\textbf{0.29}& \multicolumn{9}{c}{} \\ \cline{1-9}
\end{tabular}
\end{center}
    \caption{Average \colt{} ratings for experiment team pairs. All 95\% confidence intervals are less than 0.015.}
    \label{tab:coltx}
\end{table}
\vspace*{\fill}
\end{landscape}

\begin{landscape}
\vspace*{\fill}
\begin{table}[ht]
\begin{center}
    %Win Ratecross table
\begin{tabular}{|c|c|c|c|c|c|c|c|c|c|c|c|c|c|c|c|c|c|}
\hline
& \multicolumn{7}{c|}{\textbf{Guessers}} & \multicolumn{5}{c|}{\textbf{With Partner}} & \multicolumn{5}{c|}{\textbf{Without Partner}}\\
 \hline
\textbf{SM}& \textbf{w} \hc{}& \textbf{wg} \hc{}& \textbf{g5} \hc{}& \textbf{g1} \hc{}& \textbf{g2} \hc{}& \textbf{g3} \hc{}& \textbf{cn} \hc{}& \textbf{A} \hc{}& \textbf{B} \hc{}& \textbf{ACE} \hc{}& \textbf{BA} \hc{}& \textbf{R} \hc{}& \textbf{A} \hc{}& \textbf{B} \hc{}& \textbf{ACE} \hc{}& \textbf{BA} \hc{}& \textbf{R} \hc{}\\ \hline 
\textbf{w} \hc{}& \textbf{1.00}& 0.81& 0.62& 0.68& 0.72& 0.73& \textbf{0.83}& 0.77& 1.00\bc{}& 1.00\bw{}& 0.81& 0.82& 0.73& 0.83\bc{}& 0.81\bwo{}& 0.81\bwo{}& 0.76\\ 
 \hline 
\textbf{wg} \hc{}& \textbf{0.78}& 0.88& \textbf{0.96}& 0.88& 0.86& 0.86& 0.74& 0.85& 0.96\bc{}& 0.93\bw{}& 0.88& 0.87& 0.85 & 0.96\bc{}& 0.93\bwo{}& 0.88& 0.87\\ 
 \hline 
\textbf{g5} \hc{}& 0.65& 0.82& \textbf{1.00}& \textbf{0.85}& 0.81& 0.82& 0.63& 0.80& 1.00\bc{}& 1.00\bw{}& 0.82& 0.84& 0.77& 0.85\bc{}& 0.82\bwo{}& 0.82\bwo{}& 0.79\\ 
 \hline 
\textbf{g1} \hc{}& 0.71& 0.89& 0.88& \textbf{1.00}& \textbf{0.89}& 0.89& 0.73& 0.86& 1.00\bc{}& 1.00\bw{}& 0.89& 0.89& 0.84& 0.89\bc{}& 0.88& 0.89\bwo{}& 0.86\\ 
 \hline 
\textbf{g2} \hc{}& 0.73& 0.95& 0.83& \textbf{0.89}& \textbf{1.00}& \textbf{0.95}& 0.77& 0.87& 1.00\bc{}& 0.99\bw{}& 0.95& 0.90& 0.85& 0.95\bc{}& 0.93& 0.95\bwo{}& 0.87\\ 
 \hline 
\textbf{g3} \hc{}& 0.73& \textbf{0.98}& 0.80& 0.86& \textbf{0.92}& \textbf{1.00}& 0.77& 0.87& 1.00\bc{}& 0.99\bw{}& 0.98& 0.89& 0.84& 0.98\bc{}& 0.96& 0.98\bwo{}& 0.87\\ 
 \hline 
\textbf{cn} \hc{}& 0.76& \textbf{0.77}& 0.59& 0.65& 0.71& 0.74& \textbf{1.00}& 0.75& 1.00\bc{}& 0.99\bw{}& 0.77& 0.79& 0.70& 0.77\bc{}& 0.75& 0.77\bwo{}& 0.72\\ 
 \hline 
\multicolumn{8}{c}{\textbf{With Partner}} &&\textbf{0.99}\bcf{}&\textbf{0.99}\bwf{}&\textbf{0.87}&\textbf{0.86}&&\textbf{0.89}\bcf{}&\textbf{0.87} \bwf{} &\textbf{0.87}\bwf{}&\textbf{0.82}\\ \cline{1-8}\cline{10-13}\cline{15-18}
\textbf{A}\hc{}& 0.77& 0.87& 0.81& 0.83& 0.85& 0.86& 0.78& \multicolumn{9}{c}{} \\ \cline{1-9}
\textbf{B}\hc{}& 1.00\bc{}& 0.98\bc{}& 1.00\bc{}& 1.00\bc{}& 1.00\bc{}& 1.00\bc{}& 1.00\bc{}&\textbf{1.00}\bcf{}& \multicolumn{9}{c}{} \\ \cline{1-9}
\textbf{ACE}\hc{}& 0.99\bw{}& 0.96\bw{}& 0.99\bw{}& 0.99\bw{}& 0.99& 0.99\bw{}& 0.99\bw{}&\textbf{0.99}\bwf{}& \multicolumn{9}{c}{} \\ \cline{1-9}
\textbf{BA}\hc{}& 0.73& 0.95& 0.83& 0.89& 1.00\bw{}& 0.95& 0.77&\textbf{0.87}& \multicolumn{9}{c}{} \\ \cline{1-9}
\textbf{R}\hc{}& 0.81& 0.88& 0.86& 0.86& 0.87& 0.88& 0.82&\textbf{0.85}& \multicolumn{9}{c}{} \\ \cline{1-9}
\multicolumn{8}{c}{\textbf{Without Partner}} & \multicolumn{10}{c}{} \\ \cline{1-8}
\textbf{A} \hc{}& 0.73& 0.87& 0.78& 0.80& 0.82& 0.83& 0.74& \multicolumn{10}{c}{} \\ \cline{1-9}
\textbf{B} \hc{}& 0.78\bc{}& 0.98\bc{}& 0.96\bc{}& 0.89\bc{}& 0.92\bc{}& 0.95\bc{}& 0.83\bc{}&\textbf{0.90}\bcf{}& \multicolumn{9}{c}{} \\ \cline{1-9}
\textbf{ACE} \hc{}& 0.76\bwo{}& 0.96\bwo{}& 0.94\bwo{}& 0.87& 0.90\bwo{}& 0.93& 0.80\bwo{}&\textbf{0.88}\bwf{}& \multicolumn{9}{c}{} \\ \cline{1-9}
\textbf{BA} \hc{}& 0.73& 0.95& 0.83& 0.89\bwo{}& 0.86& 0.95\bwo{}& 0.77&\textbf{0.85}& \multicolumn{9}{c}{} \\ \cline{1-9}
\textbf{R} \hc{}& 0.75& 0.88& 0.82& 0.82& 0.84& 0.85& 0.77&\textbf{0.82}& \multicolumn{9}{c}{} \\ \cline{1-9}
\end{tabular}
\end{center}
    \caption{Average win rates for experiment team pairs. All 95\% confidence intervals are less than 0.0052.}
    \label{tab:wrx}
\end{table}
\vspace*{\fill}
\end{landscape}

\begin{landscape}
\vspace*{\fill}
\begin{table}[ht]
\begin{center}
    %Average Win Timecross table
\begin{tabular}{|c|c|c|c|c|c|c|c|c|c|c|c|c|c|c|c|c|c|}
\hline
& \multicolumn{7}{c|}{\textbf{Guessers}} & \multicolumn{5}{c|}{\textbf{With Partner}} & \multicolumn{5}{c|}{\textbf{Without Partner}}\\
 \hline
\textbf{SM}& \textbf{w} \hc{}& \textbf{wg} \hc{}& \textbf{g5} \hc{}& \textbf{g1} \hc{}& \textbf{g2} \hc{}& \textbf{g3} \hc{}& \textbf{cn} \hc{}& \textbf{A} \hc{}& \textbf{B} \hc{}& \textbf{ACE} \hc{}& \textbf{BA} \hc{}& \textbf{R} \hc{}& \textbf{A} \hc{}& \textbf{B} \hc{}& \textbf{ACE} \hc{}& \textbf{BA} \hc{}& \textbf{R} \hc{}\\ \hline 
\textbf{w} \hc{}& \textbf{3.87}& 5.68& 7.50& 6.99& 6.61& 6.46& \textbf{5.59}& 6.10& 3.87\bc{}& 3.90\bw{}& 5.68& 5.73& 6.47& 5.59\bc{}& 5.78& 5.68\bwo{}& 6.32\\ 
 \hline 
\textbf{wg} \hc{}& 6.38& 5.21& \textbf{4.45}& 5.18& 5.45& 5.42& 6.81& 5.56& 4.45\bc{}& 4.71\bw{}& 5.21& 5.39& 5.56& 4.45\bc{}& 4.71\bwo{}& 5.21& 5.39\\ 
 \hline 
\textbf{g5} \hc{}& 7.64& 5.80& \textbf{4.07}& \textbf{5.40}& 5.84& 5.86& 7.85& 6.07& 4.07\bc{}& 4.10\bw{}& 5.80& 5.72& 6.40& 5.40\bc{}& 5.78\bwo{}& 5.80& 6.19\\ 
 \hline 
\textbf{g1} \hc{}& 6.99& 5.06& 5.17& \textbf{4.09}& \textbf{5.05}& 5.07& 6.91& 5.48& 4.09\bc{}& 4.13\bw{}& 5.06& 5.23& 5.71& 5.05\bc{}& 5.23& 5.06\bwo{}& 5.51\\ 
 \hline 
\textbf{g2} \hc{}& 6.72& 4.51& 5.62& \textbf{5.08}& \textbf{4.05}& \textbf{4.48}& 6.48& 5.28& 4.05\bc{}& 4.10\bw{}& 4.51& 5.04& 5.48& 4.48\bc{}& 4.69& 4.51\bwo{}& 5.26\\ 
 \hline 
\textbf{g3} \hc{}& 6.50& \textbf{4.03}& 5.73& 5.16& \textbf{4.51}& \textbf{3.85}& 6.13& 5.13& 3.85\bc{}& 3.91\bw{}& 4.03& 4.88& 5.35& 4.03\bc{}& 4.17& 4.03\bwo{}& 5.11\\ 
 \hline 
\textbf{cn} \hc{}& \textbf{6.01}& \textbf{5.99}& 7.86& 7.19& 6.58& 6.34& \textbf{3.70}& 6.24& 3.70\bc{}& 3.75\bw{}& 5.99& 5.84& 6.66& 5.99\bc{}& 6.18& 5.99\bwo{}& 6.53\\ 
 \hline 
\multicolumn{8}{c}{\textbf{With Partner}} &&\textbf{4.01}\bcf{}&\textbf{4.09}\bwf{}&\textbf{5.18}&\textbf{5.40}&&\textbf{5.00}\bcf{}&\textbf{5.22}&\textbf{5.18}\bwf{}&\textbf{5.76}\\ \cline{1-8}\cline{10-13}\cline{15-18}
\textbf{A}\hc{}& 6.30& 5.18& 5.77& 5.59& 5.44& 5.35& 6.21& \multicolumn{9}{c}{} \\ \cline{1-9}
\textbf{B}\hc{}& 3.87\bc{}& 4.03\bc{}& 4.07\bc{}& 4.09\bc{}& 4.05\bc{}& 3.85\bc{}& 3.70\bc{}&\textbf{3.95}\bcf{}& \multicolumn{9}{c}{} \\ \cline{1-9}
\textbf{ACE}\hc{}& 3.93\bw{}& 4.21\bw{}& 4.12\bw{}& 4.14\bw{}& 4.10& 3.92\bw{}& 3.76\bw{}&\textbf{4.03}\bwf{}& \multicolumn{9}{c}{} \\ \cline{1-9}
\textbf{BA}\hc{}& 6.72& 4.51& 5.62& 5.08& 4.05\bw{}& 4.48& 6.48&\textbf{5.28}& \multicolumn{9}{c}{} \\ \cline{1-9}
\textbf{R}\hc{}& 5.82& 4.97& 5.36& 5.29& 5.16& 5.06& 5.74&\textbf{5.34}& \multicolumn{9}{c}{} \\ \cline{1-9}
\multicolumn{8}{c}{\textbf{Without Partner}} & \multicolumn{10}{c}{} \\ \cline{1-8} 
\textbf{A} \hc{}& 6.71& 5.18& 6.06& 5.83& 5.67& 5.60& 6.63& \multicolumn{10}{c}{} \\ \cline{1-8}
\textbf{B} \hc{}& 6.01\bc{}& 4.03\bc{}& 4.45\bc{}& 5.08\bc{}& 4.51\bc{}& 4.48\bc{}& 5.59\bc{}&\textbf{4.88}\bcf{}& \multicolumn{9}{c}{} \\ \cline{1-9}
\textbf{ACE} \hc{}& 6.40\bwo{}& 4.21\bwo{}& 4.60\bwo{}& 5.24& 4.84\bwo{}& 4.69& 6.04\bwo{}&\textbf{5.15}\bwf{}& \multicolumn{9}{c}{} \\ \cline{1-9}
\textbf{BA} \hc{}& 6.72& 4.51& 5.62& 5.08\bwo{}& 5.45& 4.48\bwo{}& 6.48&\textbf{5.48}& \multicolumn{9}{c}{} \\ \cline{1-9}
\textbf{R} \hc{}& 6.44& 4.97& 5.69& 5.59& 5.43& 5.35& 6.40&\textbf{5.70}& \multicolumn{9}{c}{} \\ \cline{1-9}
\end{tabular}
\end{center}
    \caption{Average win times for experiment team pairs. All 95\% confidence intervals are less than 0.03.}
    \label{tab:wtx}
\end{table}
\vspace*{\fill}
\end{landscape}
\endgroup

\subsection{Performance over Time}
We now look at how the performance of the \ace{} agent changes over time as it utilizes feedback gained by playing with a specific teammate. 
In particular, we are concerned with how many games it takes to start achieving good performance. 
The results previously discussed were averages over all 50 games in each experiment. 
But is the performance poor at the beginning of the 50 games and only starts improving towards the end? 
How quickly does the \ace{} agent adapt?

\begin{figure}
    \centering
         \begin{subfigure}[b]{0.45\textwidth}
         \centering
         \includegraphics[width=\textwidth]{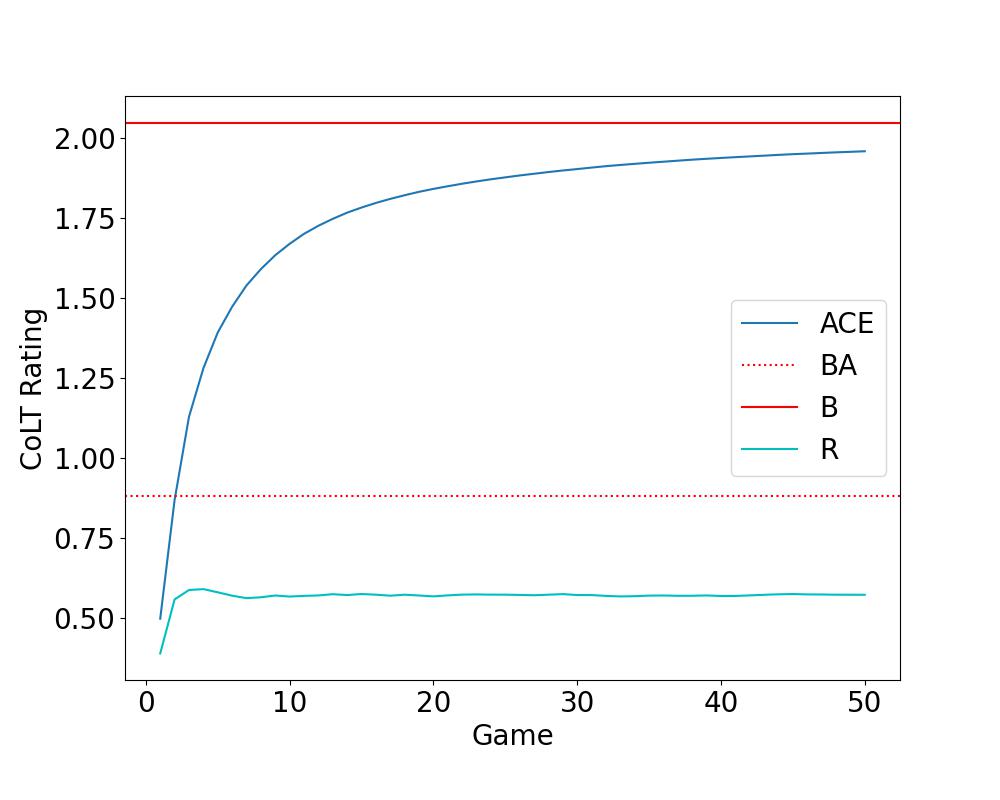}
         \caption{\ace{} Spymaster with g1 Guesser}
         \label{fig:s1}
     \end{subfigure}
     \hfill
          \begin{subfigure}[b]{0.45\textwidth}
         \centering
         \includegraphics[width=\textwidth]{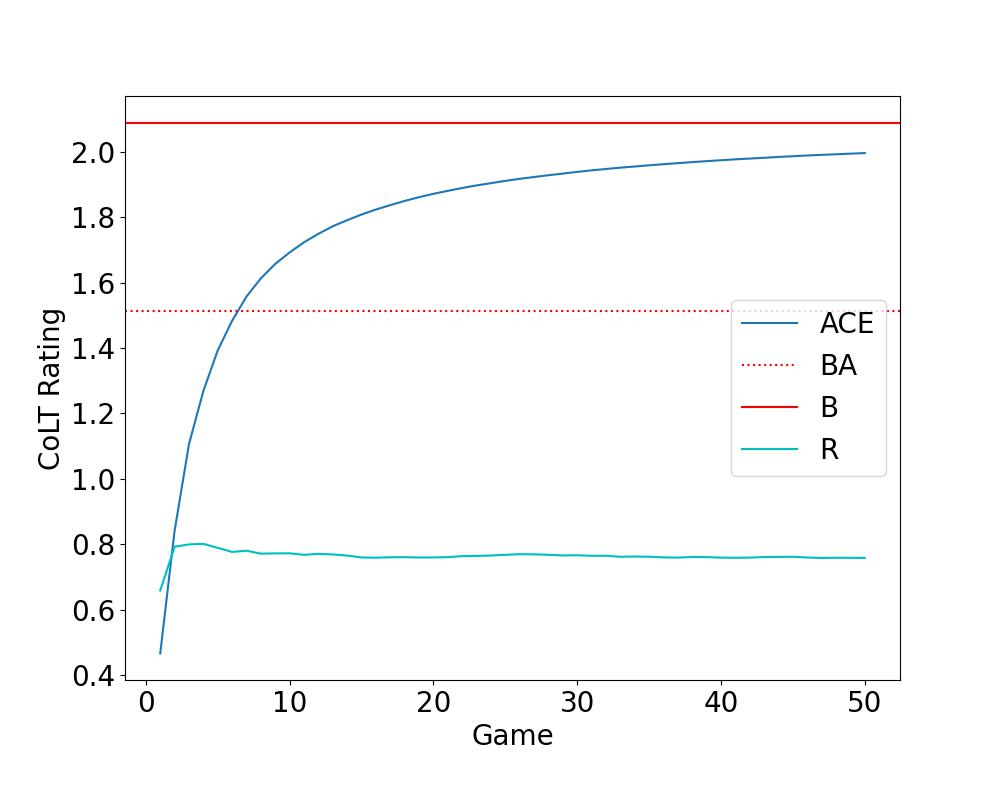}
         \caption{\ace{} Spymaster with g3 Guesser}
         \label{fig:s2}
     \end{subfigure}
     \hfill
          \begin{subfigure}[b]{0.45\textwidth}
         \centering
         \includegraphics[width=\textwidth]{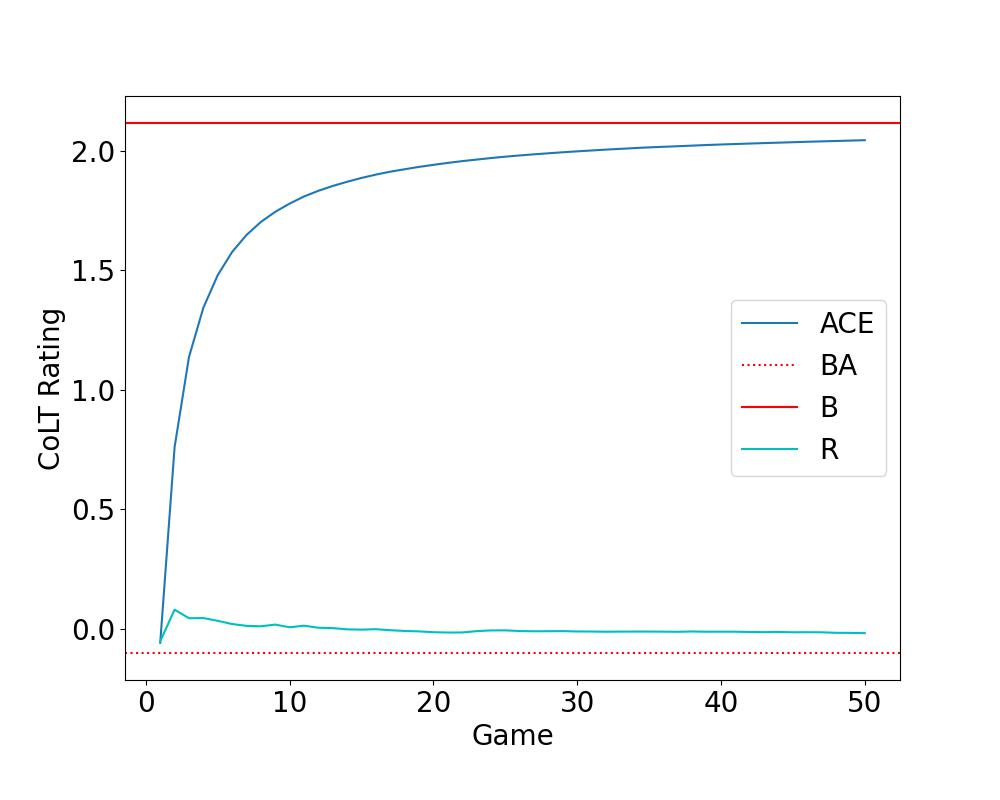}
         \caption{cn Spymaster with \ace{} Guesser}
         \label{fig:s3}
     \end{subfigure}
     \hfill
          \begin{subfigure}[b]{0.45\textwidth}
         \centering
         \includegraphics[width=\textwidth]{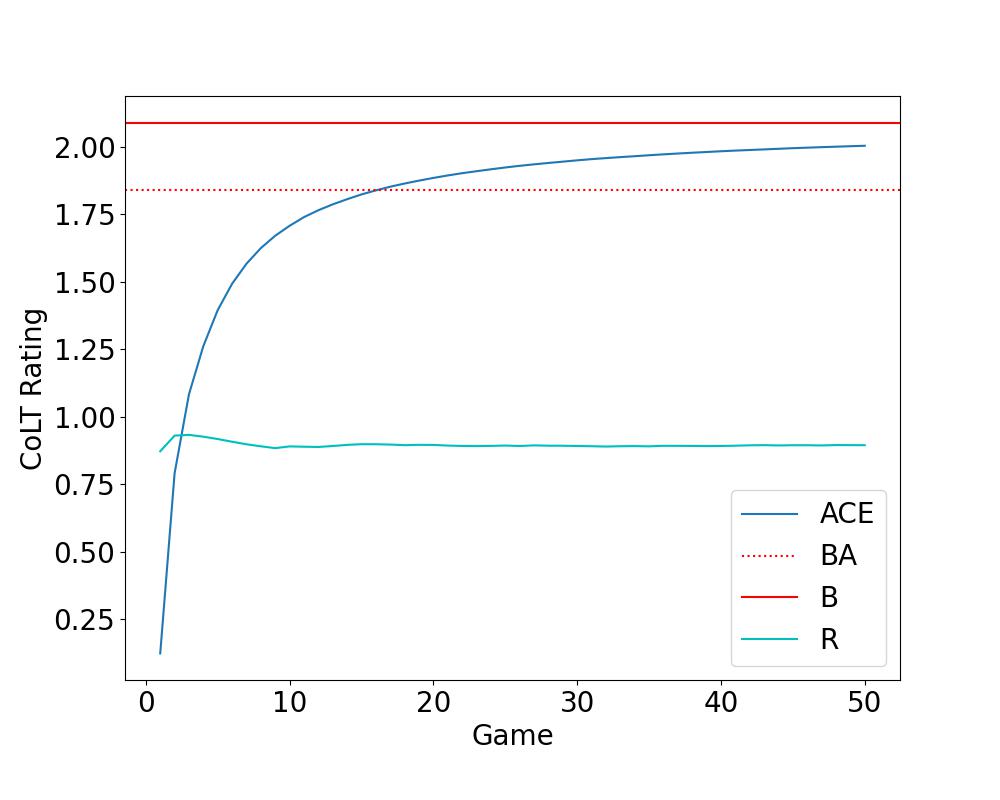}
         \caption{g3 Spymaster with \ace{} Guesser}
         \label{fig:s4}
     \end{subfigure}
    \caption{Average progress of \colt{} rating for \emph{with partner} pairings}
    \label{fig:coltprog}
\end{figure}

Figure \ref{fig:coltprog} shows a few examples the progression of the \colt{} score for some \emph{with partner} pairings involving the \ace{} agent. 
We can see that the \colt{} score for the teams converges to close to the best one as the teams interact. 
Additionally, the score rises very quickly during the first 10 games and then begins to approach the score of the best expert more asymptotically. 

Another way of viewing the performance of the \ace{} agent over time is shown in Table \ref{tab:onlyend}.  
The table shows the average \colt{} score for each pairing, when the first $t$ games, out of the 50 total in each experiment, are excluded. 
The \colt{} scores at $t=0$ are the same as were shown in Table \ref{tab:coltx}, which are the \colt{} scores taking into account all time steps.
The highest $t$ value included is $t=40$ so that there are still 10 games over which to compute the \colt{} score. 
Table \ref{tab:onlyend} shows how the \colt{} score increases, due to the \ace{} agent's improved performance over time. 
Cells are highlighted in yellow when they correspond to at least 50\% of the total \colt{} score improvement from $t=0$ to $t=40$. 

In the \emph{with partner} case, after only two games as both the spymaster and the guesser, the \ace{} agent is 66\% of the way to its final performance value. 
In the \emph{without partner} case, it takes longer for the \ace{} agent's \colt{} score to increase, but after 10 and 5 games, for \ace{} as the spymaster and guesser respectively, the average \colt{} score has increased at least 50\% of the way to its final value.  
When \ace{} is the guesser, it's average \colt{} score is at least the same as the best average (which was 0.80 in Table \ref{tab:coltx}) from game 10 onward. 

These results show that the \ace{} agent can adapt quickly to its teammate. 
When a very good partner agent is present in the ensemble, this adaptation happens extremely quickly, over the course of just a few games. 
This is a short enough time scale to be able to adapt to human players as teammates. 

\begin{table}[]
    \centering
    %Sliding Window Average Pair Score table
\begin{tabular}{|c|c|cccccccc|}
\multicolumn{10}{c}{With Partner} \\ 
 \hline 
&& \multicolumn{8}{c|}{$t$} \\ 
 \hline 
\textbf{Spymaster} \hc{} & \textbf{Guesser} \hc{}&\textbf{0} \hc{}&\textbf{1} \hc{}&\textbf{2} \hc{}&\textbf{5} \hc{}&\textbf{10} \hc{}&\textbf{15} \hc{}&\textbf{30} \hc{}&\textbf{40} \hc{}\\ 
 \hline 
\ace{} \hc{} & -- \hc{}&1.94&1.97&1.99\bw{}&2.00\bw{}&2.01\bw{}&2.02\bw{}&2.02\bw{}&2.02\bw{}\\ 
 \hline 
-- \hc{} & \ace{} \hc{}&1.91&1.94&1.95\bw{}&1.96\bw{}&1.97\bw{}&1.97\bw{}&1.97\bw{}&1.97\bw{}\\ 
 \hline 
\multicolumn{10}{c}{Without Partner} \\ 
 \hline 
&& \multicolumn{8}{c|}{$t$} \\ 
 \hline 
\textbf{Spymaster} \hc{} & \textbf{Guesser} \hc{}&\textbf{0} \hc{}&\textbf{1} \hc{}&\textbf{2} \hc{}&\textbf{5} \hc{}&\textbf{10} \hc{}&\textbf{15} \hc{}&\textbf{30} \hc{}&\textbf{40} \hc{}\\ 
 \hline 
\ace{} \hc{} & -- \hc{}&0.86&0.87&0.88&0.89&0.90\bw{}&0.90\bw{}&0.92\bw{}&0.93\bw{}\\ 
 \hline 
-- \hc{} & \ace{} \hc{}&0.76&0.78&0.78&0.79\bw{}&0.80\bw{}&0.80\bw{}&0.81\bw{}&0.82\bw{}\\ 
 \hline 
\end{tabular}

    \caption{Average \colt{} ratings for pairings, excluding first $t$ rounds}
    \label{tab:onlyend}
\end{table}

\section{Conclusions} \label{sec:conclusions}

In this article \ace{}, the first adaptive agent for the cooperative word game Codenames was presented.
This adaptive agent internally utilizes a set of experts, each a basic Codenames agent.
The novel \colt{} rating function was presented, which generates a rating for a Codenames team based on the distribution over outcomes that team achieves. 
Experiments in section \ref{sec:ex-results} demonstrated that \colt{} is effective, and that as it is optimized, win rate and win time are also positively increased.
We encourage future Codenames researchers to use the \colt{} rating to report on Codenames agent and team performance. 
The goal of the \ace{} agent algorithm is to maximize the \colt{} rating it obtains with its current teammate, and it uses the UCB algorithm to decide which expert to use on each turn to accomplish this.
Through experiments it was demonstrated that this approach is effective in practice and that the \ace{} agent is able to adapt to individual teammates, utilizing different expert agents with different partners.  
In particular it is able to achieve very high performance with a wide range of teammates, leveraging only information obtained from the game play as feedback. 
The \ace{} agent is able to adapt quickly when one of the ensemble experts is a significantly better match than the others.  
When all the experts do poorly, or the performance of the experts is more mixed with no standout, the performance and adaptation times are slower.
While the experiments didn't include every previous Codenames agent that has been proposed in the literature, nothing in the \ace{} agent is specific to any of the agents in the ensemble. 
The \ace{} agent should work effectively regardless of which agents appear in the ensemble.

Our hope is that this ensemble approach will enable more exploration of unique and novel Codenames agent designs, which can each compliment each other in an ensemble, as opposed to finding a single `magic' agent with the perfect language model that will be best with every teammate. 
Each new agent does not have to work well with every possible teammate that could be faced, but can instead be focused on being effective with a subset of them. 
New agent ideas can be added to an ensemble and if they help, they will be chosen by the \ace{} agent, and if not they won't negatively impact the overall performance over time. 

In the future we want to utilize the \ace{} agent to explore which Codenames agent designs and language models are most compatible with humans. 
We can also see how much humans vary in which expert is the best match for them. 
As the performance of the \ace{} agent will obviously vary based on the experts which make up the ensemble, we also plan to explore how the size and composition of the ensemble impacts performance. 
We can investigate ways of constructing an effective ensemble of reduced size from a larger set of candidate experts, and if it is possible to construct new experts on the fly by mixing over language models themselves. 
The \ace{} algorithm and \colt{} rating function enable the exploration of new frontiers in the goal to create AI agents that can adapt to any teammate in cooperative language settings. 

\bibliography{refs}
\bibliographystyle{plain}

\end{document}